# ScaleFormer: Revisiting the Transformer-based Backbones from a Scale-wise Perspective for Medical Image Segmentation

**Huimin Huang**[1], * **Shiao Xie**[1], ✉ **Lanfen Lin**[1], **Yutaro Iwamoto**[2],
**Xian-Hua Han**[3], ✉ **Yen-Wei Chen**[2], and **Ruofeng Tong**[1,4]
[1] Zhejiang University
[2] Ritsumeikan University
[3] Yamaguchi University
[4] Zhejiang Lab

## Abstract

Recently, a variety of vision transformers have been developed as their capability of modeling long-range dependency. In current transformer-based backbones for medical image segmentation, convolutional layers were replaced with pure transformers, or transformers were added to the deepest encoder to learn global context. However, there are mainly two challenges in a scale-wise perspective: (1) intra-scale problem: the existing methods lacked in extracting local-global cues in each scale, which may impact the signal propagation of small objects; (2) inter-scale problem: the existing methods failed to explore distinctive information from multiple scales, which may hinder the representation learning from objects with widely variable size, shape and location. To address these limitations, we propose a novel backbone, namely ScaleFormer, with two appealing designs: (1) A scale-wise intra-scale transformer is designed to couple the CNN-based local features with the transformer-based global cues in each scale, where the row-wise and column-wise global dependencies can be extracted by a lightweight Dual-Axis MSA. (2) A simple and effective spatial-aware inter-scale transformer is designed to interact among consensual regions in multiple scales, which can highlight the cross-scale dependency and resolve the complex scale variations. Experimental results on different benchmarks demonstrate that our Scale-Former outperforms the current state-of-the-art methods. The code is publicly available at: *https://github.com/ZJUGiveLab/ScaleFormer*.

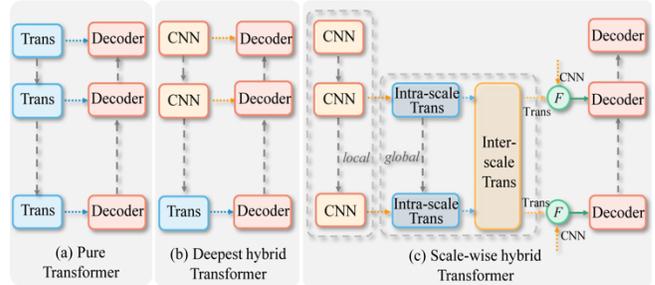

Figure 1: Comparison of current transformer-based backbones ((a) and (b)) with our ScaleFormer (c).

## 1 Introduction

In the past decades, Convolutional Neural Networks (CNNs) have greatly promoted the development of dense prediction tasks [Shelhamer *et al*., 2015], and a variety of CNN-based segmentation models have been developed [Ronneberger *et al*., 2015; Chen *et al*., 2017; Huang *et al*., 2021]. Despite its great success, CNNs failed to model explicit long-range relations beyond local regions (shown in Fig.2.(b) and (f)), since the effective receptive field of a network's units is severely limited [Luo *et al*., 2016].

Recently, the transformer architecture has been introduced to visual tasks and shown to be an efficient way of modeling global context [Dosovitskiy *et al*., 2020]. This largely attributes to the multi-head self-attention (MSA) and Multilayer Perceptron (MLP) structure. Fig.1. (a) illustrates one of the transformer-based architectures for medical image segmentation, e.g. SwinUNet [Cao *et al*., 2021] and Missformer [Huang *et al*., 2021]. They typically utilized a stack of transformer blocks to model global representations. However, these pure transformer-based architectures often failed to achieve satisfactory performance, since lack of spatial inductive-bias in modelling local information. As shown in Fig.2(c) and (g), SwinUNet activated the background extensively and was difficult to distinguish objects from background. Another popular architecture, e.g. TransUNet [Chen *et al*., 2021] and AFTer-UNet [Yan *et al*., 2022], in Fig.1(b), used a CNN-based encoder to explore detailed high-resolution spatial information, and added several transformers into the deepest layer to learn global context. However, this kind of deepest

---

✉ Corresponding Authors: Lanfen Lin (llf@zju.edu.cn), Yen-Wei Chen (chen@is.ritsumei.ac.jp).

Huimin Huang and Shiao Xie are co-first authors.

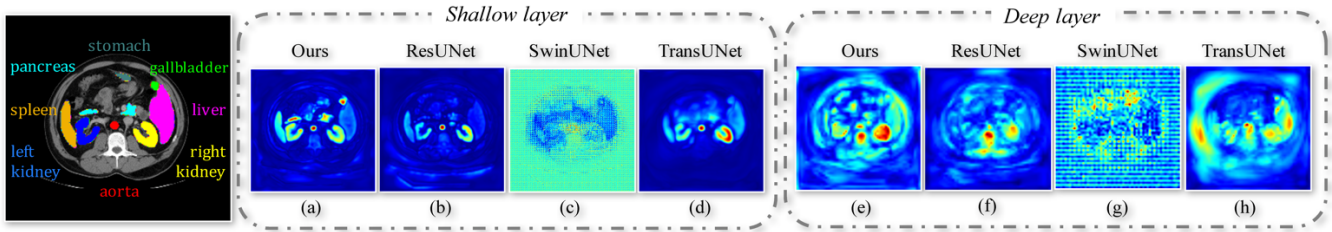

Figure 2: Visualization of feature maps from different layers of four methods. The CNN-based ResUNet (in (b), (f)) limits in modeling long-range relations and thus fails to activate on objects beyond local regions (e.g. unactivated liver, gallbladder and pancreas). The pure transformer, SwinUNet (in (c), (g)), activates background extensively as a result it is difficult to distinguish objects from background. TransUNet (in (d), (h)) stacks transformers in the deepest encoder of UNet, but it still has problems accurately localizing small objects (e.g. blurred gallbladder and pancreas with small size). Our ScaleFormer (in (a), (e)) improves the transformer-based backbone from a scale-wise perspective, which can recognize objects in various sizes and long-range scenarios. (e.g. long-distant large-size liver and small-size pancreas).

hybrid transformer structure still led to blurred and inaccurate activation on the salient objects in Fig.2.(d) and (h).

The observed unsatisfactory activations of current transformer-based backbones may be imputed to two main challenges in a scale-wise perspective. *(i) Intra-scale problem*: As we know, both contextual local-level details and positional global-level cues are essential complements in each scale, especially for small objects, whose fine signals in the shallow layer (in Fig.2(b)-(d)) may be gradually diluted during downsampling (in Fig.2 (f)-(h)). But the integration of intra-scale local features with global representations remains a fundamental challenge. *(ii) Inter-scale problem*: As shown in Fig.2, the objects in medial image often have widely variable size, shape and location. The existing methods were incapable of extracting sufficient information from multiple scales generated by hierarchical encoder, which failed to notice the complete salient objects. To address these challenges, we propose a novel backbone for medial image segmentation task (in Fig.1. (c)), namely ScaleFormer, which is motivated by the following two aspects:

Firstly, considering both the spatial difference and global distribution of objects are essential characteristics, a scale-wise intra-scale transformer is designed to couple the CNN-based local features with the transformer-based global representations in each scale. In this way, we are able to highlight both detailed spatial information (e.g. contextual cues) and long-range dependencies (e.g. positional cues). However, the quadratic computational cost of previous transformers becomes the main obstacle in such intra-scale structure, especially for the shallow scales with high-resolution feature maps. To achieve this, we also design a lightweight and universal Dual-Axis MSA to suppress less-relevant information and recognize salient parts with fast speed, which is realized by capturing row-wise and column-wise global dependencies.

Secondly, as many previous studies demonstrated, multi-scale features explore distinctive information and are capable of resolving the complex scale variations. In this inter-scale circumstance, for a specific object (e.g. liver), instead of calculating the redundant relationships with long-distant objects (e.g. spleen and aorta) at various scales, we focus on the long-range cross-scale dependency of the object itself and the nearby objects (e.g. liver and gallbladder), which can effectively capture the scale difference (e.g. shape variation) of each object, and achieve promising performance with satisfactory boundary. Inspired by this, we design a spatial-aware inter-scale transformer considering that there is a spatial correspondence among patches in various scales through downsampling operation. In this way, we focus on the cross-scale similarities among these spatial-aware patches, which efficiently learns the mutual information in a lightweight manner.

The ability of ScaleFormer in capturing local-global cues is shown in Fig.2(a), (e). By extracting the intra-scale local details and global semantics, ScaleFormer can focus on the small object in both shallow and deep layers. Additionally, benefiting from the inter-scale interactions, ScaleFormer also attends full object extent with in various sizes and long-range scenarios. The major contributions of this work are four-fold: (**i**) We analyze the intra-scale and inter-scale problems faced by the current transformer-based backbones, and propose a novel backbone in a scale-wise perspective, termed as Scale-Former, to improve the segmentation quality of medical image segmentation. (**ii**) We propose a scale-wise intra-scale transformer to associate CNN-based local features with intra-scale transformer-based global cues in each scale. Accordingly, we design a lightweight and universal Dual-Axis MSA to capture row-wise and column-wise global dependencies. (**iii**) We propose a spatial-aware inter-scale transformer to interact among consensual regions in multiple scales, which is capable of capturing cross-scale dependency and resolving complex scale variations in a simple and effective manner. (**iv**) We conduct extensive experiments on three public datasets, which shows ScaleFormer surpasses SOTA methods.

## 2 Related Work

**Vision transformers.** In the most current research, ViT [Dosovitskiy *et al*. 2020] introduced transformer into visual tasks and achieved competitive performance. Motivated by the success of ViT, SegFormer [Xie *et al.*, 2021], SETER [Zheng *et al.*, 2020], DETR [Carion *et al.,* 2020] have been proposed for semantic segmentation and object detection tasks, which achieved comparative results with CNN-based methods. To reduce the computational cost of self-attention operation in ViT, CMT [Guo *et al.*, 2021] utilized depth-wise convolution to downsample key and value to extract important information; while Swin transformer [Liu *et al*. 2021] restricted the attention in local regions. Different from existing methods, we design a novel backbone that couples local CNN blocks and global Transformer. Such a structure not only inherits sufficient structural prior from CNN but also embeds global representations from transformer block in each scale.

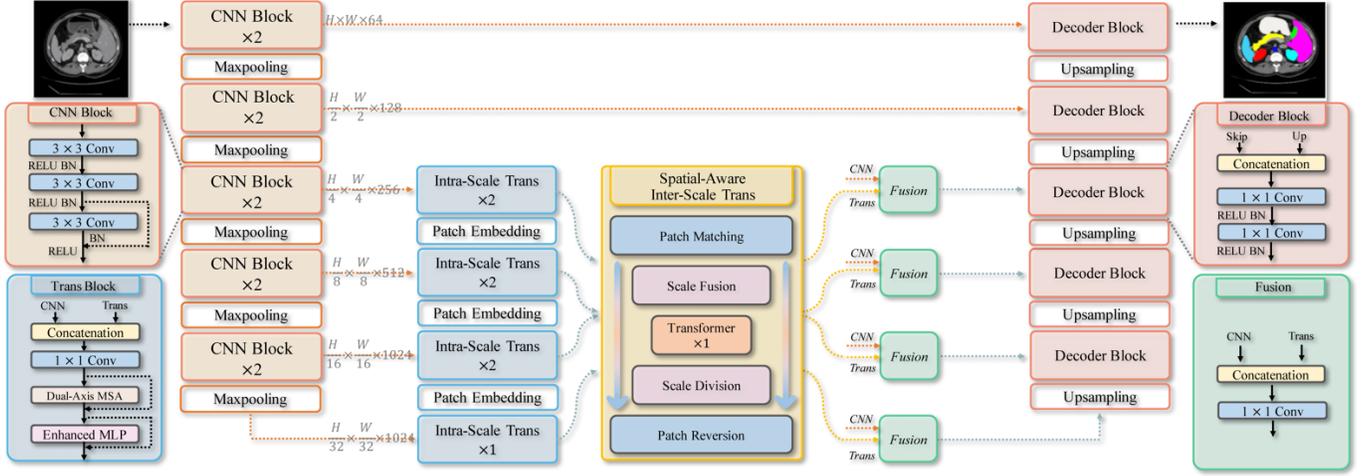

Figure 3: An overview of our ScaleFormer. First, the input image passes the CNN blocks to generate local-level details, which are sent to the intra-scale transformers to capture the global-level cues. The inter-scale transformer then takes the enhanced representation from each scale as the inputs and interacts among different scales to model the mutual information of objects. Finally, the reinforced output is further combined with the same-scale CNN features, and then send to the corresponding decoder block for the final prediction.

**Multi-scale architectures for medical image segmentation.** Multi-scale fusion has been proved to improve the segmentation performance in CNN-based methods. PSPNet [Zhao et al., 2017] exploited the capability of global context information by aggregating different region-based features. UNet++ [Zhou et al.,2020] and UNet 3+ [Huang et al., 2020] have been proposed to bridge the semantic gap between shallow encoder layers and deep decoder layers and made full use of multi-scale information through skip connection. To learn more extensive and rich context information, MS-Dual-Guided network [Sinha et al., 2020] applied attention in both spatial and channel dimension of feature maps on different scales. Considering the multi-scale cues, we further propose a simple and effective spatial-aware inter-scale transformer, which captures long-distance relationships among scales.

## 3 ScaleFormer

### 3.1 Overview

The proposed ScaleFormer learns intra-scale local-global cues as well as modeling inter-scale dependencies as shown in Fig.3. In special, we utilize ResUNet backbone to extract local features in a hierarchical manner. The encoder part has five CNN blocks comprised of basic resnet-34 blocks, where the channel number increases with network depth while the resolution of feature maps decreases. When the CNN goes deeper, different fine-grained features are extracted in various stages. Considering that transformer is powerful at capturing long-distance relationships, the scale-wise intra-scale transformer (in Sec.3.2) is coupled with the corresponding CNN block to highlight both detailed spatial information and long-range dependencies in each scale. To learn mutual regions at different scales, the spatial-aware inter-scale transformer (in Sec.3.3) further interacts among multiple scales. Finally, the enhanced representation is aggregated with the local-level CNN features in the same scale, and then sends to the decoder block for the final prediction. In the following, we will describe each step of ScaleFormer in detail.

### 3.2 Intra-Scale Transformer Block

As shown in Fig.3, the initial transformer only receives input from the CNN branch of same stage, whereas the input of other intra-scale transformer blocks incorporates global information from previous transformer stages to aggregate fine-grained details and coarse semantics information. Let $i$ indexes the down-sampling layer along transformer branch, $N_t$ refers to the total number of transformers. The stack of feature maps $X_{trans}^i$ can be represented as:

$$X_{trans}^i = \begin{cases} Trans(X_{cnn}^i), & i = 1 \\ Trans([X_{cnn}^i, Down(X_{trans}^{i-1})]), & i = 2, \cdots, N_t \end{cases} \quad (1)$$

where function $Down(\cdot)$ indicates the patch embedding mechanism, which is realized by a convolution operation with stride of 2 and followed by a batch normalization and a ReLU activation function. $[\cdot]$ is the concatenation operation. $Trans(\cdot)$ is the intra-scale transformer block, which consists of a lightweight Dual-Axis MSA module, and an enhanced MLP. We will describe these parts in the following.

**Dual-Axis MSA.** In the Fig.4, we illustrate the previous two MSA architecture and our proposed method. In original MSA [Dosovitskiy et al., 2020] (shown in Fig.4.(a)), the input $X \in \mathbb{R}^{H \times W \times C}$ is linearly transformed into query $Q$, key $K$ and value $V$ with the same shape of $HW \times C$. The computational cost of the original MSA is $O(H^2W^2C)$. To achieve lower cost, some works [Guo et al., 2021] used a spatial reduction mechanism by applying a reduction ratio $R$ to the key $K$ and value $V$ (shown in Fig.4.(b)) and the computational cost is decreased to $O(H^2W^2C/R^2)$. However, it still requires large reduction ratio $R$ when applied to high-resolution feature map, which degrades the horizontal and vertical information.

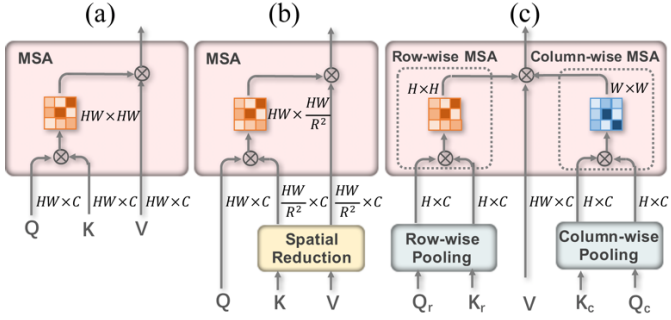

Figure 4: Comparison of previous MSAs (a) and (b) with the proposed Dual-axis MSA (c).

Different from the previous MSAs, we design a lightweight and universal Dual-Axis MSA mechanism, which replaces the full-image relation ($HW \times HW$) with a row-wise ($H \times H$) and column-wise ($W \times W$) relation, as illustrated in Fig.4.(c). To adaptively learn the horizontal and vertical representation [Chi et al., 2020], we first apply four distinctive depth-wise convolutions on $Q$ and $K$ to obtain row-wise enhanced $Q_r$, $K_r$ and column-wise enhanced $Q_c$, $K_c$. We then pass these into average pooling operation in two directions, which estimates global and compact conditions ($Q_{rp}$, $K_{rp}$, $Q_{cp}$, $K_{cp}$) over the whole feature map. The process can be formulated as:

$$A(Q,K,V) = \underbrace{\left(softmax\left(\frac{Q_{rp}K_{rp}^T}{\sqrt{d_k}}\right)\right)}_{Row-wise\ MSA} V \underbrace{\left(softmax\left(\frac{Q_{cp}K_{cp}^T}{\sqrt{d_k}}\right)\right)}_{Column-wise\ MSA} \quad (2)$$

when applying the row-wise MSA, the value $V$ is reshaped into $H \times WC$; while using the column-wise MSA, $V$ is reshaped into $HC \times W$. In this way, the complexity of our Dual-Axis MSA is $O(H^2WC + HW^2C)$. Compared with Axial Attention [Ho et al., 2019] that requires consecutive two blocks to capture long-term relations, our method learns global conditions using a single block. Then, we utilize an enhanced MLP [Huang et al., 2021] with recursive skip connection before the depth-wise convolution for feature delivery.

### 3.3 Spatial-Aware Inter-Scale Transformer Block

Multi-scale cues have been proven to be powerful in resolving complex scale variations, which are essential for medical image segmentation. Hence, we aim to design a transformer-based block to interact among different scales based on the hierarchical structure of ScaleFormer. The input sequence of enhanced intra-scale tokens is first rearranged into 2D lattice. Instead of calculating the dependencies on a long sequence that directly concatenated from multi-scale tokens (flattened from the complete 2D feature map of each scale), we take full advantage of the spatial correspondence among patches in different scales through down-sampling operation, which is called **Patch Matching**. As an example, Fig.5.(a) illustrates how to find the spatial-aware patches on three successive feature maps from *i-th* to *(i+2)-th* scale. Note that we first regularize feature maps into the same channel dimension $C$. Let denote $p_j^i$ as the *j-th* patch in the *i-th* scale, and the corresponding down-sampled patches are $p_j^{i+1}$ and $p_j^{i+2}$ sharing the bounding box with the same color (i.e. blue). In this way, we focus on the most related patches, and thus maintain the spatial correspondence as well as reduce the redundancy.

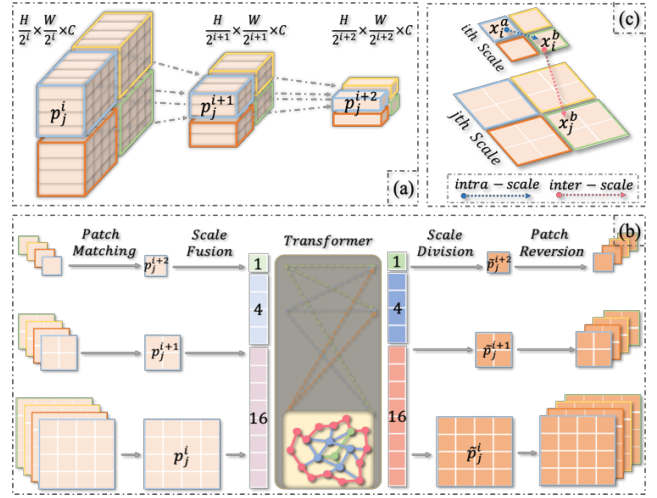

Figure 5: An illustration of spatial-aware inter-scale transformer.

Then we apply **Scale Fusion** to concatenate spatial corresponding inter-scale patches, which can be formulated as:

$$[flatten(p_j^1), \cdots, flatten(p_j^N)] \to P_j^{cat} \quad (3)$$

where $flatten(\cdot)$ operation rearrange the patch $p_j^i$ into the shape of $(HW/2^{2i}) \times C$ and $[\cdot]$ represents the concatenation operation. Because the sequence of tokens lacks horizontal and vertical information, in this part, we utilize the original MSA and MLP proposed by [Dosovitskiy et al., 2020] in our **Transformer** block, instead of using our Dual-Axis MSA (shown in eq.(2)). The spatial-aware inter-scale dependencies among patches can obtained as follows:

$$\hat{P}_j^{cat} = MSA(LN(P_j^{cat})) + P_j^{cat} \quad (4)$$

$$\tilde{P}_j^{cat} = MLP(LN(\hat{P}_j^{cat})) + \hat{P}_j^{cat} \quad (5)$$

where $LN(\cdot)$ denotes the layer normalization operator. Then we utilize **Scale Division to** reverse the enhanced sequence back to patches according to the order of concatenation:

$$Reshape\left(Split(\tilde{P}_j^{cat})\right) \to \tilde{p}_j^1, \cdots, \tilde{p}_j^N \quad (6)$$

where $Split(\cdot)$ is an inverse operation of previous concatenation operation $[\cdot]$. In the reverse procedure of patch matching,

| Method | DSC(↑) | HD(↓) | Aorta | Gallbladdr | Kidney(L) | Kidney(R) | Liver | Pancreas | Spleen | Stomach |
|---|---|---|---|---|---|---|---|---|---|---|
| V-Net | 68.81 | - | 75.34 | 51.87 | 77.10 | 80.75 | 87.84 | 40.05 | 80.56 | 56.98 |
| DARR | 69.77 | - | 74.74 | 53.77 | 72.31 | 73.24 | 94.08 | 54.18 | 89.90 | 45.96 |
| R50 U-Net | 74.68 | 36.87 | 87.74 | 63.66 | 80.60 | 78.19 | 93.74 | 56.90 | 85.87 | 74.16 |
| U-Net | 76.85 | 39.70 | 89.07 | 69.72 | 77.77 | 68.60 | 93.43 | 53.98 | 86.67 | 75.58 |
| R50 Att-UNet | 75.57 | 36.97 | 55.92 | 63.91 | 79.20 | 72.71 | 93.56 | 49.37 | 87.19 | 74.95 |
| Att-UNet | 77.77 | 36.02 | 89.55 | 68.88 | 77.98 | 71.11 | 93.57 | 58.04 | 87.30 | 75.75 |
| R50 ViT | 71.29 | 32.87 | 73.73 | 55.13 | 75.80 | 72.20 | 91.51 | 45.99 | 81.99 | 73.95 |
| TransUnet | 77.48 | 31.69 | 87.23 | 63.13 | 81.87 | 77.02 | 94.08 | 55.86 | 85.08 | 75.62 |
| SwinUNet | 79.12 | 21.55 | 85.47 | 66.53 | 83.28 | 79.61 | 94.29 | 56.58 | 90.66 | 76.60 |
| AFTer-UNet | 81.02 | - | **90.91** | 64.81 | **87.90** | **85.30** | 92.20 | 63.54 | 90.99 | 72.48 |
| MISSFormer | 81.96 | 18.20 | 86.99 | 68.65 | 85.21 | 82.00 | 94.41 | **65.67** | **91.92** | **80.81** |
| ScaleFormer(w/o inter) | 82.08 | 18.03 | 88.29 | **76.48** | 83.10 | 80.59 | 94.88 | 63.90 | 89.29 | 80.12 |
| ScaleFormer | **82.86** | **16.81** | 88.73 | 74.97 | 86.36 | 83.31 | **95.12** | 64.85 | 89.40 | 80.14 |

Table 1: Comparison to state-of-the-art (SOTA) methods on Synapse dataset.

*patch reversion* assembles all patches in the same scale into a feature map. Then we concatenate the global-level transformer feature and the local-level CNN feature as skip connection, which further sends to the decoder stage.

**Discussion.** In this section, we analyze the successive operations of the intra-scale and inter-scale transformers. The local features firstly passed through intra-scale transformer to extract the contextual information, which learns the entire pixel-wise dependency in each scale. Then, the enhanced feature maps are fed into inter-scale transformer, which explores the corresponding patch-wise dependencies among multiple scales. In this way, even though two pixels ($x_i^a$, $x_j^b$) in different scales ($i$ and $j$) and irrelevant patches ($a$ and $b$), in Fig5.(c), their relationship can be also captured via first finding the intra-scale dependencies of $x_i^a$ and $x_i^b$, then calculating the inter-scale dependencies of $x_i^b$ and $x_j^b$. These two consecutive steps realize the goal of capturing pixel-wise dependencies in full scales with less computational cost.

## 4 Experiments

### 4.1 Dataset and Evaluation

We perform experiments on three public medical image datasets:

**Synapse.** The multi-organ segmentation challenge [Landman et al., 2015] consists 30 abdominal CT scans. Following [Chen et al., 2021], we split 18 cases for training and remaining 12 cases for testing. We reported the Dice Coefficient (DSC) and Hausdorff Distance (HD) on 8 different organs.

**MoNuSeg.** The Multi-Organ Nucleus Segmentation Challenge [Kumar et al., 2017] contains 30 images for training, and 14 images for testing. Following [Wang et al., 2021], we utilized DSC and Intersection over Union (IoU) metrics.

**ACDC.** The automated cardiac diagnosis challenge [Bernard et al., 2018] contains 100 MRI scans with three organs, left ventricle (LV), right ventricle (RV) and myocardium (MYO). Following [Chen et al., 2021], we reported the DSC with a random split of 70 training cases, 10 validation cases and 20 testing cases. We split the dataset randomly because the previous division was unavailable.

### 4.2 Implementation Details

We utilized data augmentations to avoid overfitting, including random rotation and flipping. Note that our ScaleFormer did not require pretrain on any large datasets. Here, we itemized the batch size (bs), learning rate (lr), maximum training epochs (ep), optimizer (opt) for three datasets:

- Synapse: bs=8; lr=3e-3; ep=600; opt=SGD;
- ACDC: bs=8; lr=3e-3; ep=200; opt=SGD;
- MoNuSeg: bs=4; lr=1e-3; ep=200; opt=Adam;

All models were trained with momentum 0.9 and weight decay 1e-4. For fair comparison, we used the same settings and combined cross entropy loss and dice loss for all experiments.

### 4.3 Comparison with the State-of-the-Arts

To demonstrate the effectiveness of ScaleFormer, we first compare it with 11 state-of- the-art methods on the Synapse dataset, including V-Net [Milletari et al., 2016], DARR [Fu et al., 2020], U-Net, Att-UNet [Schlemper et al., 2019], ViT, TransUnet, SwinUNet, AFTer-UNet, MISSFormer. Both TransUnet and SwinUNet are pretrained on the ImageNet. Here, we abbreviate ResNet-50 as "R50", which is utilized as encoder and combined with U-Net, Att-UNet and ViT.

Experimental results are reported in Table 1 and the best results are highlighted in bold. As shown, our ScaleFormer without inter-scale former (w/o inter) outperforms all the other methods in both regional measures DSC (82.08%) and boundary-aware measure HD (18.03mm). In particular, our ScaleFormer (w/o inter) surpasses other transformer-based methods, including SwinUNet and MISSFormer with pure transformers, as well as TransUNet and AFTer-UNet that stack several transformers in the deep layers of encoder. It indicates that the structure combined with CNN and intra-scale transformer has the ability of capturing local-global clues. Additionally, an increment of (DSC: +0.78% and HD: -1.22mm) is achieved when utilizing inter-scale transformer, which achieves the new SOTA on synapse dataset.

| Methods | DSC(↑) | IOU(↑) |
|---|---|---|
| UNet | 73.97 | 59.42 |
| UNet++ | 75.28 | 60.89 |
| AttUNet | 76.20 | 62.64 |
| MRUNet | 77.54 | 63.80 |
| MedT | 79.24 | 65.73 |
| TransUNet | 79.20 | 65.68 |
| SwinUNet | 78.49 | 64.72 |
| UCTransNet | 79.87 | 66.68 |
| ScaleFormer(w/o inter) | 79.59 | 66.27 |
| ScaleFormer | **80.06** | **66.87** |

Table 2: Comparison to SOTA methods on MoNuSeg dataset.

| Methods | DSC(↑) | RV | Myo | LV |
|---|---|---|---|---|
| R50 U-Net | 87.55 | 87.10 | 80.63 | 94.92 |
| R50 Att-UNet | 86.75 | 87.58 | 79.20 | 93.47 |
| R50 ViT | 87.57 | 86.07 | 81.88 | 94.75 |
| TransUNet | 89.71 | 88.86 | 84.53 | 95.73 |
| SwinUNet | 90.00 | 88.55 | 85.62 | 95.83 |
| TransUNet* | 89.63 | 86.70 | 86.96 | **95.33** |
| SwinUNet* | 89.16 | 87.04 | 86.22 | 94.24 |
| ScaleFormer(w/o inter) | 89.74 | 86.54 | 87.66 | 95.01 |
| ScaleFormer | **90.17** | **87.33** | **88.16** | 95.04 |

Table 3: Comparison to SOTA methods on ACDC dataset.

To demonstrate the generality of ScaleFormer, we conduct experiments on another two public datasets. Additionally, we compare it with UNet++, MRUNet [Ibtehaz *et al.*,2020], MedT [Valanarasu *et al.*,2021] and UCTransNet [Wang *et al.*, 2021], which are four SOTA methods on MoNuSeg dataset. Table 2 presents DSC and IoU metrics on MoNuSeg, and Table 3 reports DSC on our divided ACDC dataset (*) with retrained TransUnet and SwinUnet using their original settings. Our ScaleFormer achieves the best result for capturing more useful representations, which demonstrates the superior generalization and robustness of ScaleFormer.

Fig.6 shows the qualitative comparison of classic UNet, TransUNet, SwinUNet and our ScaleFormer on Synapse dataset. It can be observed that ScaleFormer not only accurately localizes organs but also produces coherent boundaries, even in small object circumstances.

### 4.4 Ablation Study

**Ablation of intra-scale transformers on different stages.** We investigate how intra-scale transformers affect performance among different scales on synapse dataset, which is varied from Stage6 (deepest layer with size of 7×7) to Stage2 (shallowest layer with size of 112×112). As shown in Fig.7, the accuracy is improved with stacking the transformer blocks, which achieves the highest results (DSC: 82.08%, HD:18.03mm) when extracting intra-scale local-global cues on four consecutive stages (Stage6543). However, applying the transformer on the Stage2 may affect the genuine characteristics and increase the computational cost (required two GPUs). Therefore, in our experiments, we introduce the intra-scale transformers on Stage3-Stage6, which provides the best results within a considerable computational complexity.

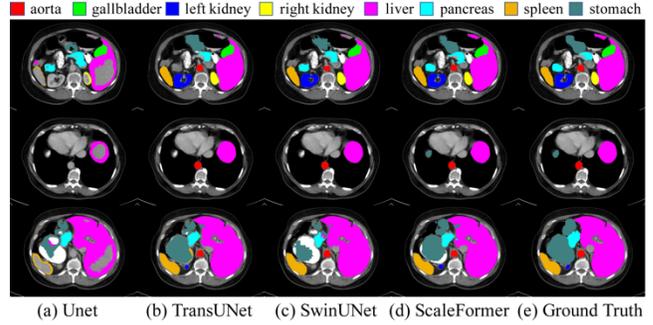

Figure 6: Qualitative results of different models on Synapse dataset.

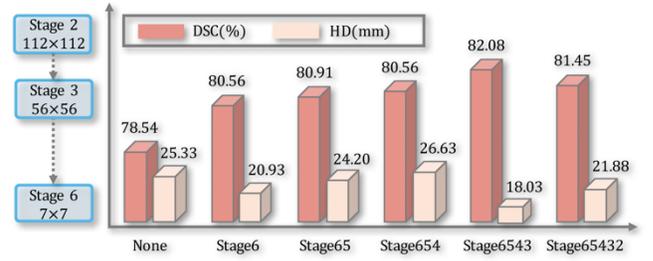

Figure 7: Ablation of intra-scale transformers on different stages

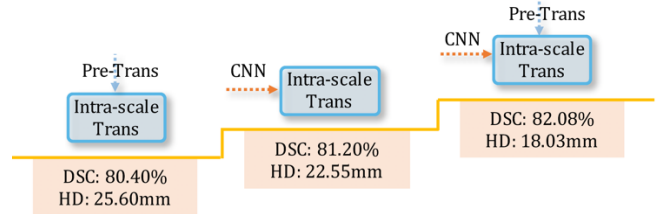

Figure 8: Ablation of the input for intra-scale transformer block.

**Ablation of the input for intra-scale transformer block.** Based on the architecture with four-stage intra-scale transformers, we present the ablation of its input, i.e. the same-stage CNN and the previous-stage transformer (Pre-Trans). As shown in Fig.8 that only using pre-trans in a staking manner achieves inferior results than using CNN feature (DSC: -0.80%, HD: +3.05mm), which means that staking transformers may dilute detailed signals by extracting global information. By combining two inputs, the best results (DSC: 82.08%, HD: 18.03mm) are achieved for aggregating fine-grained details and coarse-grained semantics.

**Effectiveness of our Dual-Axis MSA.** To verify the capability of Dual-Axis MSA in capturing global cues, we compare it with CMT [Guo *et al.*, 2021], PVT [Wang *et al.* 2021], and Axial attention [Ho *et al.*, 2019] in Fig.9(b). We find that our method achieves a better accuracy (higher DSC and lower HD) than these methods with less computation cost and fewer parameters. It indicates the superiority of Dual-Axis MSA via splitting the full-image dependency in the row-wise and the column-wise manner.

**Ablation of inter-scale transformers on various stages.** Additionally, we conduct several experiments to assess the influence of inter-scale transformers on various stages. As shown in Fig.9 (a), by learning interactive information on deeper stages, an increment from 81.37% to 82.86% is achieved, where 0.67% and 0.82% improvements are produced by adding Stage5 and Stage6. It indicates that the learned mutual relations can enhance segmentation quality.

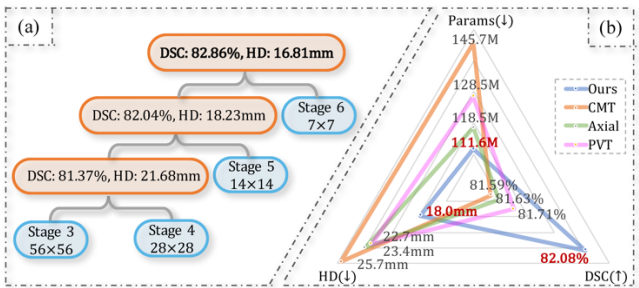

Figure 9: (a) Ablation of inter-scale transformers on various stages; (b) Comparison of Dual-Axis MSA with CMT, PVT and Axial attention on Params, DSC and HD.

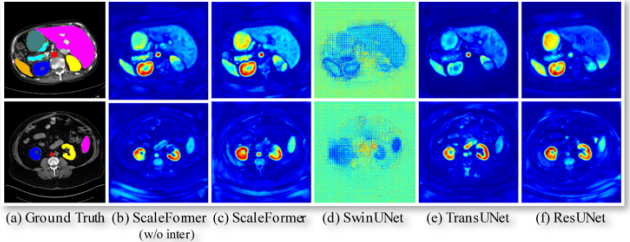

Figure 10: Visualization of the feature map of various methods.

### 4.5 Interpretation of ScaleFormer

We visualize the feature map from the final decoder in Fig.10. As shown, ResUNet (in Fig.10.(f)) has the capability of focusing on the local areas but it fails to activate larger regions; while TransUNet (in Fig.10.(e)) and SwinUNet (in Fig.10.(d)) can globally model the long-term relationships but decline the important detailed local features. Impressively, our ScaleFormer (w/o inter) (in Fig.10.(b)) not only inherits sufficient structural prior from CNN but also embeds the global representations from transformer block, which helps to find the tiny objects. Considering the multi-scale cues, ScaleFormer (in Fig.10.(c)) has more complete attention areas on objects with widely variable size, shape and location.

## 5 Conclusion

In this paper, we proposed a novel transformer-based backbone, namely ScaleFormer, from a scale-wise perspective to improve the medical image segmentation quality even for the small objects. In each scale, a scale-wise intra-scale transformer with a lightweight Dual-Axis MSA is designed to combine the CNN-based local features and transformer-based global features; while among multiple scales, a simple and effective spatial-aware inter-scale transformer was proposed to interact across scales. Experiments showed that ScaleFormer outperformed both CNN-based and transformer-based segmentation networks on three datasets.

## Acknowledgements

This work was supported in part by Zhejiang Provincial Natural Science Foundation of China under the Grant No. LZ22F020012, Major Scientific Research Project of Zhejiang Lab under the Grant No. 2020ND8AD01, and in part by the Grant-in Aid for Scientific Research from the Japanese Ministry for Education, Science, Culture and Sports (MEXT) under the Grant No. 20KK0234, No. 21H03470 and No. 20K21821.